%% file: report.tex
\documentclass[letterpaper, 10 pt, journal, twoside]{IEEEtran}

\usepackage{graphicx}
\usepackage{algorithm}
\usepackage{algorithmicx}
\usepackage[noend]{algpseudocode}
\usepackage{cite}
\usepackage{color}
\usepackage[dvipsnames]{xcolor}
\usepackage{amsmath,amssymb,amsfonts,dsfont}
\usepackage{bm}
\usepackage{setspace}
\usepackage{multirow}
\usepackage{mathtools}
\usepackage{fancyhdr}
\usepackage{authblk}
\usepackage{url}
\usepackage{tikz}

\usepackage[
  separate-uncertainty = true,
  multi-part-units = repeat
]{siunitx}
\usepackage{threeparttable}

\usepackage{renew}      

\usepackage[switch]{lineno} 
\usepackage{amsmath}   
\usepackage{background}
\usepackage{stackengine}
\setstackgap{L}{6pt}
\SetBgContents{\normalsize\stackunder[40.5cm]{\Longstack{%
{This article has been accepted for publication in a future issue of this journal, but has not been fully edited. Content may change prior to final publication. Citation information: DOI 10.1109/LRA.2018.2852777, IEEE Robotics} \\ 
 {and Automation Letters}}}{
    \rotatebox{0}{2377-3766 (c) 2018 IEEE. Personal use is permitted, but republication/redistribution requires IEEE permission. See \url{http://www.ieee.org/publications_standards/publications/rights/index.html} for more information.}}}
\SetBgScale{0.65}
\SetBgAngle{0}
\SetBgOpacity{1}
\SetBgColor{black}
\SetBgPosition{current page.south west}
\SetBgHshift{16.5cm}
\SetBgVshift{22cm}		

\begin{document}
\title{
Real-world Multi-object, Multi-grasp Detection
}

\author{Fu-Jen Chu, Ruinian Xu and Patricio A. Vela
\thanks{Manuscript received: February 24, 2018; Revised May 15, 2018; Accepted June 14, 2018.}
\thanks{This paper was recommended for publication by Editor Han Ding upon evaluation of the Associate Editor and Reviewers' comments.
This work was supported in part by NSF Award \#1605228.} 
\thanks{Fu-Jen Chu, Ruinian Xu and Patricio A. Vela are with Institute for Robotics and Intelligent Machines, 
    Georgia Institute of Technology, GA, USA.
    {\tt\footnotesize \{fujenchu, gtsimonxu, pvela\}@gatech.edu}}%
\thanks{Digital Object Identifier (DOI): see top of this page.}
}

\markboth{IEEE Robotics and Automation Letters. Preprint Version. Accepted June, 2018}
{Chu \MakeLowercase{\textit{et al.}}: Real-world Multi-object, Multi-grasp Detection} 



\maketitle

\begin{abstract}
A deep learning architecture is proposed to predict graspable locations for
robotic manipulation. It considers situations where no, one, or multiple 
object(s) are seen. By defining the learning problem to be classification
with null hypothesis competition instead of regression, the deep neural
network with RGB-D image input predicts multiple grasp candidates for a
single object or multiple objects, in a single shot. 
The method outperforms state-of-the-art approaches on the Cornell dataset
with 96.0\% and 96.1\% accuracy on image-wise and object-wise splits,
respectively. 
Evaluation on a multi-object dataset illustrates the generalization
capability of the architecture.
Grasping experiments achieve 96.0\% grasp localization and 
89.0\% grasping success rates on a test set of household objects.  
The real-time process takes less than .25\,s from image to plan.
\end{abstract}
\IEEEpeerreviewmaketitle

\section{Introduction}
\input{intro.tex}

\section{Related Work}
\input{review.tex}

\section{Problem Statement}
\input{problem.tex}

\section{Approach}
\input{approach.tex}

\section{Experiments and Evaluation}

\input{experiment.tex}

\section{Conclusion}
\input{conc.tex}

\bibliographystyle{IEEEtran}
\ifCLASSOPTIONcaptionsoff
  \newpage
\fi
\bibliography{regular,crowdsourcing}

\end{document}

%% file: intro.tex
\IEEEPARstart{W}{hile} manipulating objects is relatively easy for humans, reliably grasping
arbitrary objects remains an open challenge for robots.  Resolving it
would advance the application of robotics to industrial use cases, such
as part assembly, binning, and sorting. Likewise, it would
advance the area of assistive robotics, where the robot interacts with its
surroundings in support of human needs.  Robotic grasping involves
perception, planning, and control. As a starting point, knowing which
object to grab and how to do so are essential aspects.  
Consequently, accurate and diverse detection of robotic grasp candidates
for target objects should lead to a better grasp path planning and
improve the overall performance of grasp-based manipulation tasks.

The proposed solution utilizes a deep learning strategy for identifying
suitable grasp configurations from an input image.  
In the past decade, deep learning has achieved major success on detection,
classification, and regression tasks
\cite{bo2013unsupervised,krizhevsky2012imagenet,ngiam2011multimodal}.
Its key strength is the ability to leverage large quantities of labelled and
unlabelled data to learn powerful representations without hand-engineering
the feature space. 
Deep neural networks have been shown to outperform hand-designed
features and reach state-of-the-art performance. 
\begin{figure}[t]
    \centering
    \includegraphics[width=0.5\columnwidth]{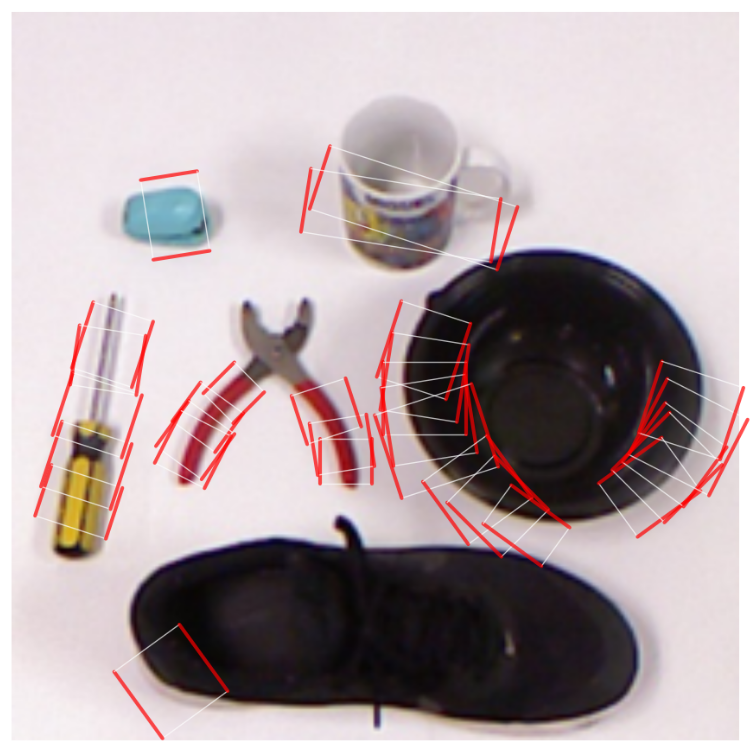}
    \caption{Simultaneous multi-object, multi-grasp detection by the proposed 
    model. Training used the Cornell dataset with the standard object-wise 
    split. The red lines correspond to parallel plates of the grasping gripper.
    The white lines indicate the distance between the plates before the
    grasp is executed.}
    \figlabel{rectangle}
\end{figure}

In this research problem, we are interested in tackling the problem of
identifying viable candidate robotic grasps of objects in a RGB-D image. 
The envisioned gripper is a parallel plate gripper 
(or similar in functionality).
The principal difficulty comes from the variable shapes and poses of 
objects as imaged by a camera.
The structural features defining successful grasps for each object may be
different; all possible features should be identified through the
learning process.
The proposed architecture associated to the grasp configuration estimation
problem relies on the strengths of deep convolutional neural networks
(CNNs) at detection and classification. Within this architecture,
the identification of grasp configuration for objects is broken down into a
grasp detection processes followed by a more refined grasp orientation
classification process, both embedded within two coupled networks.  


The proposed architecture includes a grasp region proposal network for
identification of potential grasp regions.
The network then partitions the grasp configuration estimation problem into
regression over the bounding box parameters, and classification of the
orientation angles, from RGB-D data. Importantly, the orientation classifier
also includes a {\em No Orientation} competing class to reject
spuriously identified regions for which no single orientation
classification performs well, and to act as a competing no-grasp class. 
The proposed approach predicts grasp candidates in more realistic
situations where no, single, and multiple objects may be visible; it
also predicts multiple grasps with confidence
scores (see Fig. \figref{rectangle}, scores not shown for clarity). 
A new multi-objects grasp dataset is
collected for evaluation with the traditional performance metric of
false-positives-per-image.  We show how the multi-grasp output and
confidence scores can inform a subsequent planning process to take
advantage of the multiple outputs for improved grasping success.

%


The main contributions of this paper are threefold: \\
{\bf (1)} A deep network architecture that predicts multiple
grasp candidates in situations when none, single or multiple objects are
in the view. 
Compared to baseline methods, the classification-based
approach demonstrates improved outcomes on the Cornell dataset \cite{cornell2013}
benchmark, achieving state-of-the-art performance on image-wise and
object-wise splits. \\
{\bf (2)} A multi-object, multi-grasp dataset is collected and manually
annotated with grasp configuration ground-truth as the
Cornell dataset. We demonstrate the generalization capabilities of the
architecture and its prediction performance on the multi-grasp dataset
with respect to false grasp candidates per image versus grasp miss rate.  
The dataset is available at \url{github.com/ivalab/grasp_multiObject}. \\
{\bf (3)} Experiments with a 7 degree of freedom manipulator and a
time-of-flight RGB-D sensor quantify the system's ability to grasp a
variety of household objects placed at random locations and
orientations.  Comparison to published works shows that the approach is
effective, achieving a sensible balance for real-time object pick-up
with an 89\% success rate and less than 0.25\,s from image to prediction to
plan. 

%
%
%

%% file: review.tex
Research on grasping has evolved significantly over the last two
decades.  Altogether, the review papers 
\cite{shimoga1996robot,bicchi2000robotic,sahbani2012overview,bohg2014data} 
provide a good context for the overall field.  This section reviews
grasping with an emphasis on learning-based approaches and on
representation learning.


Early work on perception-based learning approaches to grasping goes back
to \cite{kamon1996learning}, which identified a low dimensional feature
space for identifying and ranking grasps. Importantly it showed that
learning-based methods could generalize to novel objects.  Since then,
machine learning methods have evolved alongside grasping strategies.
Exploiting the input/output learning properties of machine learning (ML)
systems, \cite{saxena2008robotic} proposed to learn the image to grasp
mapping through the manual design of convolutional
networks. As an end-to-end system, reconstruction of the object's 3D
geometry is not needed to arrive at a grasp hypothesis.  The system was
trained using synthetic imagery, then demonstrated successful grasping
on real objects.  Likewise, \cite{asif2017rgb} employed a CNN-like
feature space with random forests for grasp identification.
In addition to where to grasp,
several efforts learn the grasp approach or pre-grasp strategy
\cite{ekvall2007learning,huebner2008selection}, while some focus on whether the
hypothesized grasp is likely to succeed \cite{le2010learning}.  Many of
these approaches exploited contemporary machine learning algorithms with
manually defined feature spaces.

At the turn of this decade, two advances led to new methods for
improving grasp identification: the introduction of low-cost depth
cameras, and the advent of computational frameworks to facilitate the
construction and training of CNNs.
The advent of consumer depth cameras enabled models of grasping mapping
to encode richer features.
In particular \cite{jiang2011efficient} represented grasps as a 2D
oriented rectangle in the image space with the local surface normal as
the approaching vector; this grasp configuration vector has been adopted
as the well-accepted formulation.  
Generally, the early methods using depth cameras sought to recover the
3D geometry from point clouds for grasp planning \cite{rao2010grasping},
with manually derived feature space used in the learning process. Deep
learning approaches arrived later \cite{krizhevsky2012imagenet} and were 
quickly adopted by the computer vision community \cite{he2016deep}.

Deep learning avoids the need for engineering feature spaces, with the
trade-off that larger datasets are needed.  The trade-off is usually
mitigated through the use of pre-training on pre-existing computer vision
datasets followed by fine-tuning on a smaller, problem-specific dataset.
Following a sliding window approach, \cite{lenz2015deep} trained a two
stage multi-modal network, with the first stage generating hypotheses for a
more accurate second stage. Similarly,
\cite{wang2016robot} first performed an image-wide pre-processing step to
identify candidate object regions, followed by application of a CNN classifier for each region.
To avoid sliding windows or image-wide search, end-to-end approaches are
trained to output a single grasp configuration from the input data
\cite{redmon2015real,kumra2016robotic,watson2017real}.
Regression-based approaches \cite{redmon2015real} require compensation
through image partitioning since the grasp configuration space is
non-convex.
Following the architecture in \cite{redmon2015real}, \cite{watson2017real} 
experimented on real objects with physical grasping experiments.
The two-stage network in \cite{kumra2016robotic} first output a learnt
feature, which was then used to provide a single grasp output. These approaches may
suffer from averaging effects associated to the single-output nature of
the mapping.  Guo et al. \cite{GuEtAl_ICRA2017} employed a two-stage
process with a feature learning CNN, followed by specific deep network
branches (graspable, bounding box, and orientation). 
Alternatively, \cite{johns2016deep, MaEtAl_RSS[2017]} predicted grasp
scores over the set of all possible grasps (which may be discretized).
Such networks admit the inclusion of end-effector uncertainty. 
Most deep network approaches mentioned above start with
the strong prior that every image contains a single object with a single
grasp target (except \cite{wang2016robot}).  This assumption does not
generically hold in practice: many objects have multiple grasp options
and a scene may contain more than one object. 

Another line of research is to learn the mapping from vision input to robot motion to achieve grasping. To directly plan grasps, Lu et al. \cite{lu2017planning} proposed to predict and maximize grasp success by inferring grasp configurations from
vision input for grasp planning. Research on empirical grasp planning 
with reinforcement learning (RL) acquired samples from robots in real
experiments \cite{pinto2016supersizing}. The training time involved
several weeks and led to limitation of its scalability.  The work 
\cite{levine2016learning} collected over 800k data points with up to 14
robotic arms running in parallel for learning visual servoing. 
The training time involved over 2 months. Generalization performance of
RL solution to environmental changes remains unknown.


Inspired by \cite{ren2015faster}, we propose to incorporate a
\emph{grasp region proposal network} to generate candidate regions for feature
extraction.
Furthermore, we propose to transform grasp configuration from a
regression problem formulated in previous works \cite{redmon2015real,
kumra2016robotic} into a combination of region detection and orientation
classification problems (with null hypothesis competition). 
We utilize ResNet \cite{he2016deep}, the current state-of-the-art deep
convolutional neural network, for feature extraction and grasp
prediction.
Compared to previous approaches, our method considers more realistic
scenarios with multiple objects in a scene. The proposed architecture
predicts multiple grasps with corresponding confidence scores, which
aids the subsequent planning process and actual grasping.

%% file: problem.tex
\begin{figure}
  \vspace*{0.07in}
  \hspace*{-0.02in}
  \centering
  \begin{tikzpicture} [outer sep=0pt, inner sep=0pt]
  \scope[nodes={inner sep=0,outer sep=0}] 
  \node[anchor=east] (a) 
    {\includegraphics[width=4cm,clip=true,trim=0.5in 0.25in 1.0in 0.35in]{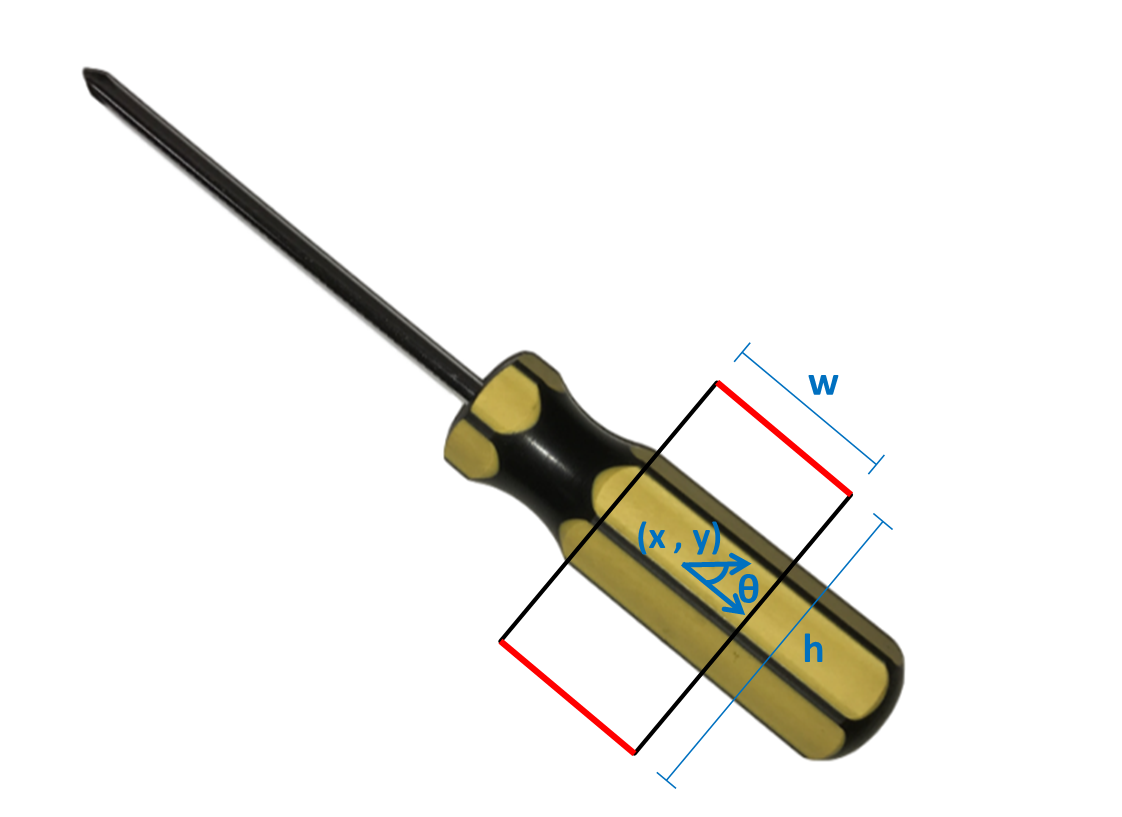}}; 
  \node[anchor=south west] (b) 
    {\includegraphics[width=2.5cm]{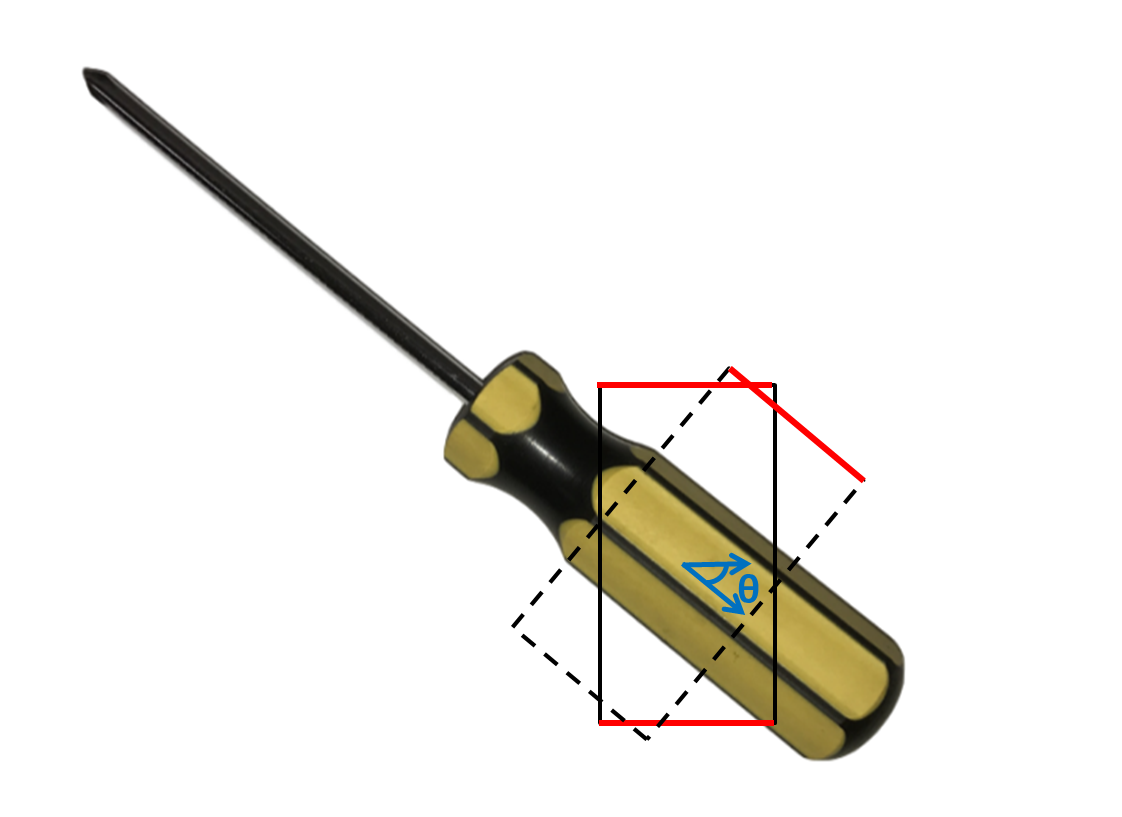}}; 
  \node[anchor=north west] (c) 
    {{\includegraphics[width=2.5cm,clip=true,trim=0in 0.3in 0in 0in]{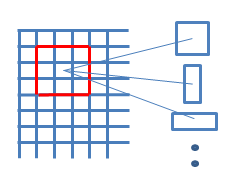}}}; 
  \endscope
  \foreach \n in {a,b,c} { 
    \node[anchor=north east] at (\n.north west) {(\n)};
  }
  \end{tikzpicture}
  \vspace*{-1.0ex}
  \caption{
    (a) The 5D grasp representation. 
    (b) A grasp rectangle is first set to the zero orientation for grasp 
    proposal training. The angle $\theta$ is one of the discrete rotation
    angles. 
    (c) Each element in the feature map is an anchor and corresponds
    to multiple candidate grasp proposal bounding boxes.
    \label{fig_rectangle}}
  \vspace*{-1.0ex}
\end{figure}

Given corresponding RGB and depth images of a novel object, the
objective is to identify the grasp configurations for potential grasp
candidates of an object for the purpose of manipulation.  
The 5-dimensional \emph{grasp rectangle} is the grasp representation employed
\cite{jiang2011efficient}.
It is a simplification of the 7-dimensional representation 
\cite{jiang2011efficient} and describes the location, orientation, and
opening distance of a parallel plate gripper prior to closing on an object. 
The 2D orientated rectangle, shown in Fig. \ref{fig_rectangle}a depicts
the gripper's location $(x,y)$, orientation $\theta$, and opening distance
$h$.  An additional parameter describing the length $w$ completes the
bounding box grasp configuration,
\begin{align} \label{eqn_grasp_representation}
  g = {\{x,y,\theta,w,h\}^T}.
\end{align}

Thinking of the center of the bounding box with its local $(x,y)$ axes
aligned to the $w$ and $h$ variables, respectively, the first three parameters
represent the $SE(2)$ frame of the bounding box in the image, while the last
two describe the dimensions of the box.


%% file: approach.tex

Much like \cite{redmon2015real,GuEtAl_ICRA2017}, the proposed approach
should avoid sliding window such as in \cite{lenz2015deep} for real-time
implementation purposes. We avoid the time-consuming sliding-window
approach by harnessing the capacity of neural networks to perform
bounding box regression, and thereby to predict candidate regions on the
full image directly.  Furthermore, we preserve all possible grasps and
output all ranked candidates, instead of regressing a single outcome.
To induce a richer feature representation and learn more of the
structural cues, we propose to use a deeper network model compared to
previous works \cite{redmon2015real, lenz2015deep, pinto2016supersizing},
with the aim of improving feature extraction for robotic grasp detection. 
We adopt the ResNet-50 \cite{he2016deep} with 50 layers, which has more
capacity and should learn better than the AlexNet
\cite{krizhevsky2012imagenet} used in previous works (8 layers). 
ResNet is known for its residual learning concept to overcome the challenge
of learning mapping functions. A residual block is designed as an incorporation
of a skip connection with standard convolutional neural network. This design
allows the block to bypass the input, and encourage convolutional layers to
predict the residual for the final mapping function of a residual block. 

The next three subsections describe the overall architecture of the
system.  It includes integration of the proposal network with a
candidate grasp region generator; a description of our choice to define
grasp parameter estimation as a combination of regression and
classification problems; and an explanation of the multi-grasp detection
architecture.    

\subsection{Grasp Proposals}

%

The first stage of the deep network aims to generate grasp proposals across
the whole image, avoiding the need for a separate object segmentation
pipeline. 

\begin{figure*}[t]
    \centering
    \hspace*{-0.08in}
    \fbox{\includegraphics[width=1\textwidth]{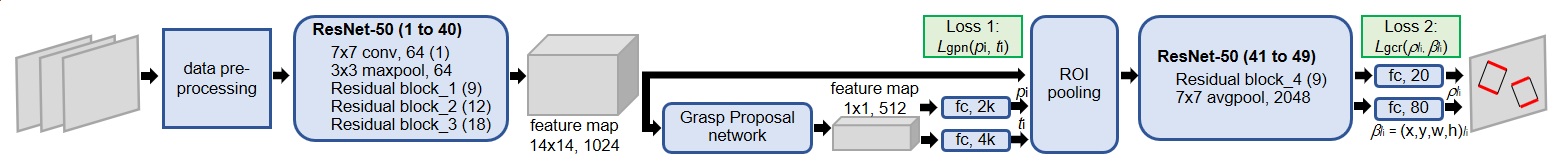}}
    
    \caption{Complete structure of our multi-object multi-grasp predictor.
    The network takes RG-D inputs, and predicts multiple grasps
    candidates with orientations and rectangle bounding boxes for each
    object in the view. Blue blocks indicate network layers and gray blocks
    indicate images and feature maps. Green blocks show the two loss
    functions. The grasp proposal network slides across anchors of
    intermediate feature maps from ResNet-50 with $k=3 \times 3$ candidates predicted
    per anchor. The black lines of output bounding boxes denote the
    open length of a two-fingered gripper, while the red lines denote the
    parallel plates of the gripper. 
    \label{fig_structure}}
\end{figure*}

Inspired by \emph{Region Proposal Network} (RPN) \cite{ren2015faster}, 
the \emph{Grasp Proposal Network} in our architecture (Fig. \ref{fig_structure}) works as RPN and shares a common
feature map (14$\times$14$\times$1024 feature map) of intermediate convolutional layers from ResNet-50 (layer 40). The \emph{Grasp Proposal Network} outputs a
1$\times$1$\times$512 feature which is then fed into two sibling fully connected layers. The
two outputs specify both probability of grasp proposal and proposal bounding
box for each of $r$ anchors on the shared feature map. The ROI layer extracts features with corresponding proposal bounding boxes and sends to the rest of the networks.

The \emph{Grasp Proposal Network} works as sliding a mini-network
over the feature map.  
At each anchor of the feature map, by default 3 scales and 3 aspect ratios are used for grasp
reset bounding box shape variations, as shown in Fig \ref{fig_rectangle}c.  Hence $r \times 3 \times 3$ predictions would be generated in total. For ground truth, we reset each orientated ground truth bounding box to have vertical height and
horizontal width, as shown in Fig. \ref{fig_rectangle}b. Let $t_i$ denote the
4-dimensional vector specifying the reset $(x,y,w,h)$ of the $i$-th grasp
configuration, and $p_i$ denote the probability of the $i$-th grasp
proposal. 
For the index set of all proposals $\mathbf{I}$,
we define the loss of grasp proposal net (gpn) to be: 
\begin{multline}
  L_{gpn}(\{(p_i,t_i)_{i=1}^\mathbf{I}\}) 
    = \sum_i L_{gp\_cls}(p_i, p_i^{\ast})  \\
      + \lambda \sum_i p_i^{\ast} L_{gp\_reg}(t_i, t_i^{\ast}). 
\end{multline}
%
%
where $L_{gp\_cls}$ is the cross entropy loss of grasp proposal
classification (gp\_cls), $L_{gp\_reg}$ is the $l_1$ regression loss of
grasp proposal (gp\_reg) with weight $\lambda$. 
We denote $p_i^{\ast} = 0$ for no grasp and $p_i^{\ast} = 1$ when
a grasp is specified. The variable $t_i^{\ast}$ is the ground truth grasp
coordinate corresponding to $p_i^{\ast}$. 

Compared to the widely applied selective search used in R-CNN
\cite{girshick2014rich}, RPN learns object proposals end-to-end from the
input without generating region of interests beforehand.  This latter,
streamlined approach is more applicable to real-time robotic applications.  

\subsection{Grasp Orientation as Classification}
Many prior approaches \cite{kumra2016robotic, redmon2015real} regress to a
single 5-dimensional grasp representation $g={\{x,y,w,h,\theta\}}$ for a
RGB-D input image. Yet to predict either on $SE(2)$ (planar pose) or on $S^1$
(orientation) involves predicting coordinates that lies in a non-Euclidean
(non-convex) space where regression and its standard L2 loss may not perform
well.  
Rather than performing regression, our multi-grasp localization pipeline
quantizes the grasp representation orientation coordinate $\theta$ into $R$ 
equal-length intervals (each interval is represented by its centroid), and
formulates the input/ouput mapping as a classification task for grasp
orientation.  
It differs from \cite{GuEtAl_ICRA2017} in that we add a non-grasp collecting
orientation class for explicit competition with a null hypotheis.  
If none of the orientation classifiers outputs a score
higher than the non-grasp class, then the grasp proposal is considered
incorrect and rejected.  In contrast, \cite{GuEtAl_ICRA2017} has a
separate grasp confidence score, which may not capture well the
orientation-dependent properties of grasps.
The value of the non-grasp class is that it is necessary for the
downstream multi-object, multi-grasp component of the final algorithm.
The total number of classes is $\left | \mathbf{C}
\right | = R + 1$. 
Denote by $\{(l_i,\theta_i)\}_{i=1}^\mathbf{I}$ where the $i$-th grasp
configuration with classification label $l_i \in {1, . . . , R}$ is
associated with the angle $\theta_i$.  For the case of no possible
orientation (i.e., the region is not graspable), the output label is $l = 0$
and there is no associated orientation. In this paper, $R = 19$ is utilized. 

%
%


\subsection{Multi-Grasp Detection}
After the region proposal stage of the deep network, the last stage
identifies candidate grasp configurations. This last stage classifies the
predicted region proposals from previous stage into $R$ regions for grasp
configuration parameter $\theta$. At the same time the last stage also
refines the proposal bounding box to a non-oriented grasp bounding box
$(x,y,w,h)$. 


To process the region proposals efficiently, we integrate an
\emph{ROI pooling layer} \cite{girshick2015fast} into ResNet-50 so that it
may share ResNet's convolutional layers.  Sharing the feature map with
previous layers avoids re-computation of features within the region of
interest. An \emph{ROI pooling layer} stacks all of the features of the
identified grasp proposals, which then get fed to two sibling fully
connected layers for orientation parameter classification $l$ and bounding
box regression $(x,y,w,h)$. The ROI pooling layer receives its input from
the intermediate convolutional layer of ResNet-50 (layer 40). 

Let $\rho_l$ denote the probability of class $l$ after a softmax layer, and
$\beta_l$
denote the corresponding predicted grasp bounding box. Define the loss
function of the grasp configuration prediction (gcr) to be:
\begin{multline}
  L_{gcr}(\{(\rho_l, \beta_l)\}_{c=0}^\mathbf{C}) 
    = \sum_c L_{gcr\_cls}(\rho_l) \\ 
      + \lambda_2 \sum_c \mathbf{1}_{c\neq 0}(c) L_{gcr\_reg}(\beta_c,
      \beta_c^\ast). 
\end{multline}
where $L_{gcr\_cls}$ is the cross entropy loss of grasp angle classification
(gcr\_cls), $L_{gcr\_reg}$ is the $l_1$ regression loss of grasp bounding boxes (gcr\_reg) with weight $\lambda_2$, and $\beta_c^{\ast}$ is the ground truth grasp bounding box.
   
With the modified ResNet-50 model, end-to-end training for grasp detection
and grasp parameter estimation employs the total loss: 
\begin{equation}
  L_{total} = L_{gpn} + L_{gcr}.  
\end{equation}
The streamlined system generates grasp proposals at the ROI layer,
stacks all ROIs using the shared feature, and the additional neurons of the
two sibling layers output grasp bounding boxes and orientations, or reject
the proposal.

%% file: experiment.tex
Evaluation of the grasp identification algorithm utilizes the Cornell
Dataset for benchmarking against other state-of-the-art algorithms. To demonstrate
the multi-object, multi-grasp capabilities, a new dataset is carefully collected and manually annotated.
Both datasets consist of color and depth images for multiple
modalities. In practice, not all possible grasps are covered by the labelled
ground truth, yet the grasp rectangles are comprehensive and representative for
diverse examples of good candidates. The scoring criteria takes into
account the potential sparsity of the grasp configuration by including an
acceptable proximity radius to the ground truth grasp configuration.

\paragraph{Cornell Dataset}
The Cornell Dataset \cite{cornell2013} consists of 885 images of 244
different objects, with several images taken of each object in various
orientations or poses. Each distinct image is labelled with multiple
ground truth grasps corresponding to possible ways to grab the object. 

\paragraph{Multi-Object Dataset}
Since the Cornell Dataset scenarios consist of one object in one image,
we collect a Multi-Object Dataset for the evaluation of the
multi-object/multi-grasp case. Our dataset is meant for evaluation and
consists of 96 images with 3-5 different objects in a single image. We
follow the same protocol as the Cornell Dataset by taking several images of
each set of objects in various orientations or poses. Multiple ground truth
grasps for each object in each image are annotated using the same
configuration definition.

\subsection{Cornell Data Preprocessing}
To reuse the pre-trained weights of ResNet-50 on COCO-2014 dataset
\cite{lin2014microsoft}, the Cornell dataset is preprocessed to fit the
input format of the ResNet-50 network. For comparison purposes, we follow
the same procedure in \cite{redmon2015real} and substitute the blue channel 
with the depth channel.  Since RGB data lies between 0 to 255, the depth
information is normalized to the same range.  The mean image is chosen
to be 144, while the pixels on the depth image with no information were
replaced with zeros.  For the data preparation, we perform extensive data
augmentation. First, the images are center cropped to obtain a 351x351
region. Then the cropped image is randomly rotated between 0 to 360
degree and center cropped to 321x321 in size.  The rotated image is
randomly translated in x and y direction by up to 50 pixels. The
preprocessing generates 1000 augmented data for each image. Finally the
image is resized to 227x227 to fit the input of ResNet-50 architecture.

\subsection{Pre-Training}
To avoid over-fitting and precondition the learning process,  we start
with the pretrained ResNet-50. As shown in Fig \ref{fig_structure}, we
implement grasp proposal layer after the third residual block by sharing
the feature map. The proposals are then sent to the ROI layer and fed into
the fourth residual layer. The $7\times7$ average pool outputs are then
fed to two fully connected layers for final classification and
regression. All new layers beyond ResNet-50 are trained from scratch. 

Because the orientation is specified as a class label and assigned a
specific orientation, the Cornell dataset needs to be converted into the
expected output format of the proposed network.  We equally divide 180
degrees into $R$ regions (due to symmetry of the gripper)
and assign the continuous ground truth orientation to the nearest discrete
orientation.

\subsection{Training}
For training, we train the whole network end-to-end for 5 epochs on a
single nVidia Titan-X (Maxwell architecture). The initial learning rate
is set to 0.0001. And we divide the learning rate by 10 every 10000
iterations. Tensorflow is the implementation framework with
cudnn-5.1.10 and cuda-8.0 packages.  The code will be publicly released.

\subsection{Evaluation Metric}
Accuracy evaluation of the grasp parameters involves checking for proximity
to the ground truth according to established criteria \cite{redmon2015real}. 
A candidate grasp configuration is reported correct if both:
\begin{enumerate}
  \item the difference of angle between predicted grasp $g_p$ and ground
    truth $g_t$ is within 30 $^\circ$, and
  \item the Jaccard index of the predicted grasp $g_p$ and the ground truth
    $g_t$ is greater than 0.25, e.g.,
    \begin{equation}
    \label{eqn_jaccard_index}
    J(g_p,g_t)=\frac{\vert g_p \cap g_t \vert}{\vert g_p \cup g_t \vert} >
    0.25 \tag{5}
    \end{equation}
\end{enumerate}
The Jaccard index is similar to the Intersection over Union (IoU) threshold
for object detection.  

\section{Results}

\subsection{Single-object Single-grasp}
Testing of the proposed architecture on the Cornell Dataset, and
comparison with prior works lead to Table \tabref{sosg}.  
For this single-object/single-grasp test, the highest output score of
all grasp candidates output is chosen as the final output.
The proposed architecture outperforms all competitive methods. On image-wise
split, our architecture reaches 96.0\% accuracy; on object-wise split for
unseen objects, 96.1\% accuracy is achieved.  
We also tested our proposed architecture by replacing ResNet-50 with VGG-16
architecture, a smaller deep net with 16 layers. With VGG-16, our model
still outperforms competitive approaches. Yet the deeper ResNet-50 achieve
4.4\% more on unseen objects.
Furthermore, we experiment on RGB images without depth information with ResNet-50 version and both image-wise and object-wise split perform slightly worse than our proposed approach, indicating the effectiveness of depth.
The third column contains the run-time of methods that have reported it, as
well as the runtime of the proposed method.  Computationally, our
architecture detects and localize multiple grasps in 0.120s, which is around
8 fps and is close to usable in real time applications.  The VGG-16
architecture doubles the speed with some prediction accuracy loss.

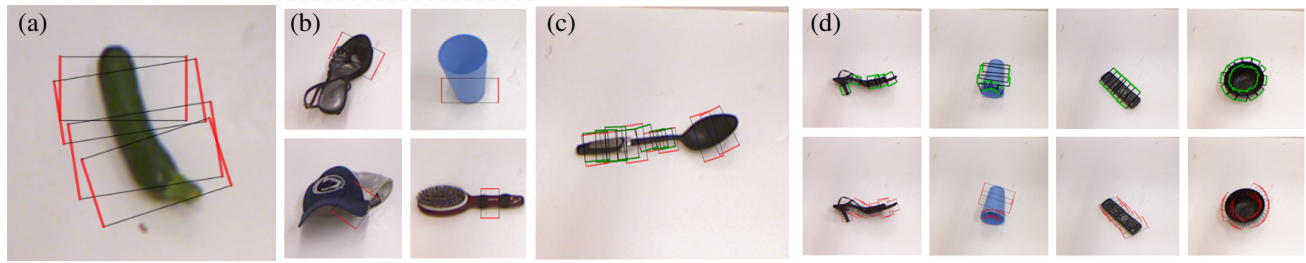
\begin{figure*}[t]
  \centering
  \vspace*{0.07in}
  \begin{tikzpicture}[inner sep=0pt]
  \node[anchor=south west] (MG) at (0in,0in) 
    {{\includegraphics[height=1.34in]{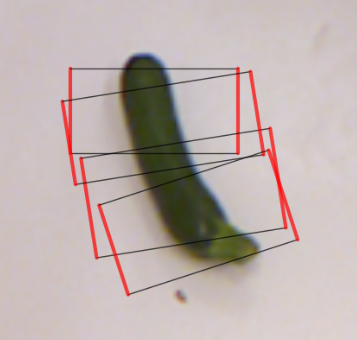}}};
  \node[anchor=south west] (SO) at (MG.south east)
    {\includegraphics[height=1.34in,clip=yes,trim=0in 0.100in 0in 0.100in]{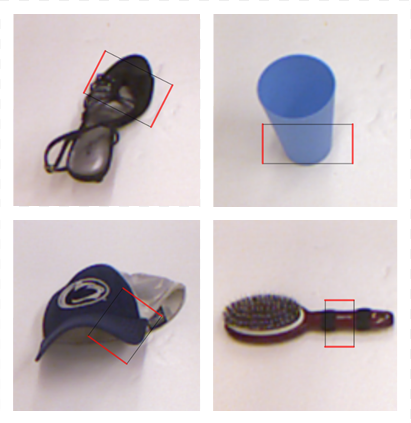}};
  \node[anchor=south west] (EG) at (SO.south east)
    {\includegraphics[height=1.34in]{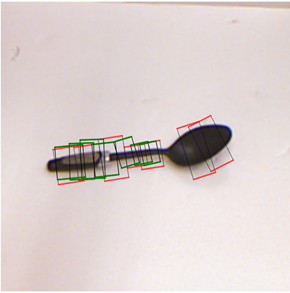}};
  \node[anchor=south west, ,xshift=4pt] (SS) at (EG.south east)
    {\includegraphics[height=1.34in,clip=true,trim=0in 4.26in 0in 0in]{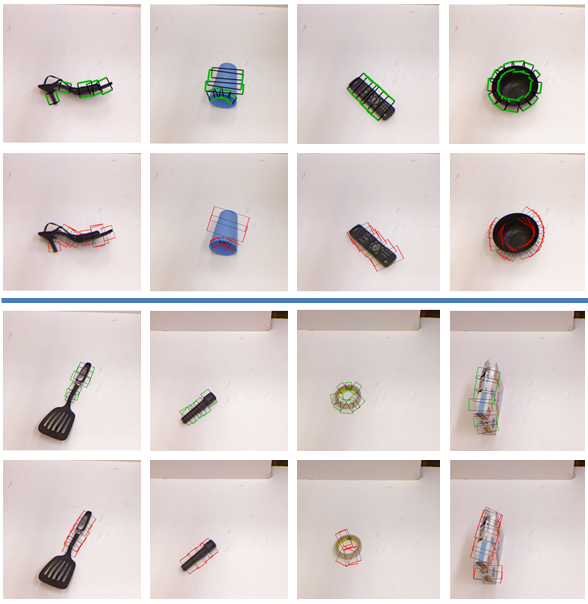}};  

  \node[anchor=north west,xshift=5pt,yshift=-3pt] (TB) at (SO.north west) {(b)};
  \node[xshift=10pt] at (TB -| MG.north west) {(a)};
  \node[xshift=10pt] (TC) at (TB -| EG.north west) {(c)};
  \node[xshift=100pt] at (TC) {(d)};
  \end{tikzpicture}
  \caption{Output 5D grasp configuration of system for Cornell dataset inputs: 
    (a) the multiple grasp options output for an object; 
    (b) the top grasp outputs for several objects; 
    (c) output grasps (red) and ground-truth grasps
    (green) showing that the system may output grasps for which
    there is no ground truth;
    (d) multi-grasp output for several objects. The green rectangles are
    ground truth and the red rectangles represent predicted grasps for 
    each unseen object.
    \label{fig_newArea} \label{fig_example_w_label} \label{singleGrasp}}
  \vspace*{-0.125in}
\end{figure*}

Table \tabref{Jthresh} contains the outcomes of stricter Jaccard indexes for
the ResNet-50 model. Performance decreases with stricter conditions but maintains competitiveness even at 0.40 IoU condition.
Typical output of the system is given in Fig. \ref{fig_example_w_label}a,
where four grasps are identified.  Limiting the output to a single grasp
leads to the outputs depicted in Fig.  \ref{singleGrasp}b. 
In the multi-grasp case, our system not only predicts universal grasps
learned from ground truth, but also contains candidate grasps not contained 
in the ground truth, Fig. \ref{fig_newArea}c.

\begin {table}[t]
  \centering
  \caption {Single-Object Single-Grasp Evaluation \tablabel{sosg} }
  \small
  \begin{tabular}{ | l | c | c | c |}
    \hline
    {\bf approach} & \bf{image-wise} & \bf{object-wise} & \bf{speed} \\ \hline
                  & \multicolumn{2}{c|}{Prediction Accuracy (\%)} & fps \\ \hline
    
    Jiang et al. \cite{jiang2011efficient}  & 60.5 & 58.3 & 0.02  \\ \hline
    Lenz et al.  \cite{lenz2015deep}        & 73.9 & 75.6 & 0.07    \\ \hline
    Redmon et al. \cite{redmon2015real}     & 88.0 & 87.1 & 3.31  \\ \hline
    Wang et al. \cite{wang2016robot}	    & 81.8 & N/A  & 7.10  \\ \hline
    Asif et al.	\cite{asif2017rgb}          & 88.2 & 87.5 & --    \\ \hline
    Kumra et al.	\cite{kumra2016robotic} & 89.2 & 88.9 & 16.03 \\ \hline
    Mahler et al. \cite{MaEtAl_RSS[2017]}   & 93.0 & N/A  & $\sim$1.25  \\ \hline
    Guo et al. \cite{GuEtAl_ICRA2017}     & 93.2 & 89.1 & --    \\
    \hline \hline 
    Ours: VGG-16	 (RGB-D)                       & \bf{95.5} & \bf{91.7} & 17.24 \\ 
    \hline 
    Ours: Res-50 (RGB)                   & \bf{94.4} & \bf{95.5} & 8.33 \\ 
    \hline 
    Ours: Res-50 (RGB-D)                        & \bf{96.0} & \bf{96.1} & 8.33 \\ 

    \hline
  \end{tabular}
  \vspace*{0.25ex}
  \caption {Prediction Accuracy (\%) at Different Jaccard Thresholds 
  \tablabel{Jthresh}}
  \begin{tabular}{ | l | c | c | c | c| }
    \hline
    \bf{split} & 0.25 & 0.30 & 0.35 & 0.40 \\ \hline
    
    \bf{image-wise}   &  \bf{96.0}  &  94.9  & 92.1  & 84.7 \\ \hline
    \bf{object-wise}  &  \bf{96.1}  &  92.7  & 87.6  & 82.6 \\ 
    \hline
  \end{tabular}
  \vspace*{-0.075in}
\end {table}

\subsection{Single-object Multi-grasp}
\begin{figure*}[t]
  \centering
  \begin{tikzpicture}[inner sep=0pt,outer sep=0pt]
    \node (SF) at (0in,0in)
      {\includegraphics[height=1.49in]{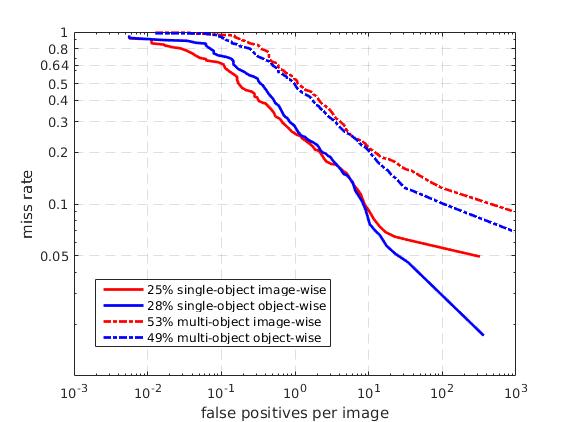}};
    \node[anchor=south west] (MF) at (SF.south east)
      {\includegraphics[height=1.49in,clip=true,trim=0in 1.3in 0in 0in]{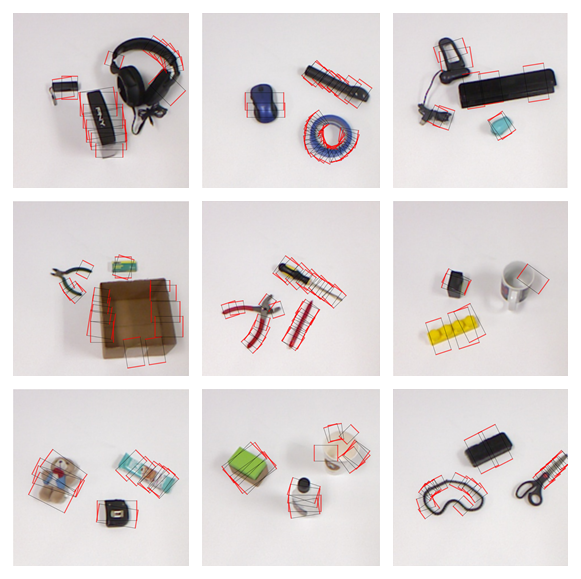}};
      
    \node[anchor=north west] (MFF) at (MF.north east)
      {\includegraphics[height=0.75in,clip=true,trim=1.35in 0.05in 1.35in 2.51in]{fig_multigrasp.png}};
    \node[anchor=south west] (MFFF) at (MF.south east)
      {\includegraphics[height=0.75in,clip=true,trim=2.6in 0.04in 0in 2.51in]{fig_multigrasp.png}};
      
    \node[anchor=south west, xshift = 0.5cm, yshift = 0.1cm] (SS) at (MFFF.south east)
     {\includegraphics[height=1.42in,clip=true,trim=0in 0in 0in 0in]{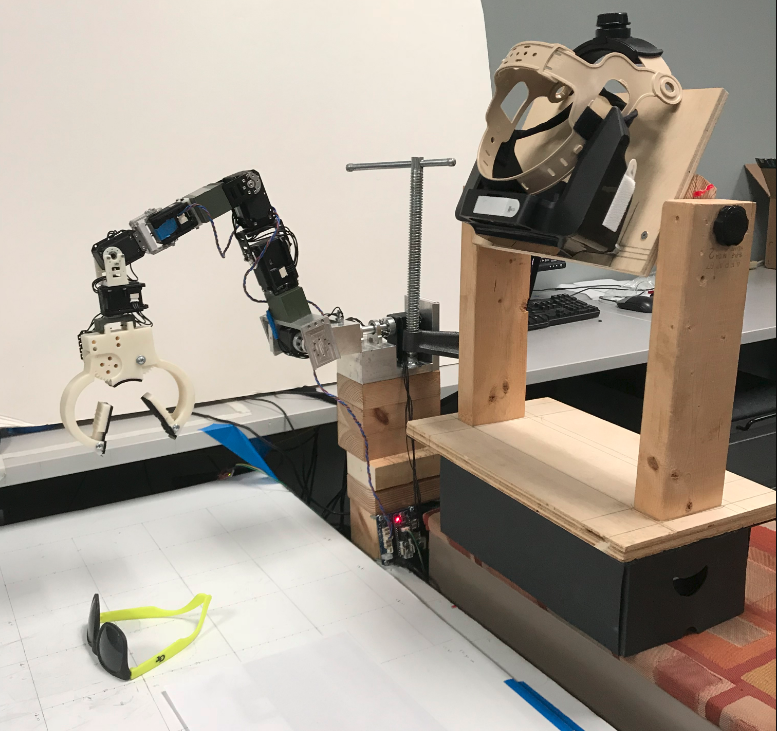}};
    \node[anchor=north west,xshift=2pt,yshift=-2pt] at (SF.north west) {(a)};
    \node[anchor=north west,xshift=5pt,yshift=-5pt] at (MF.north west) {(b)};
    \node[anchor=north west,xshift=2pt,yshift=-2pt] at (SS.north west) {(c)};
  \end{tikzpicture}
    \caption{Detection results of the system: 
    (a) The ROC curves of our system on single-object multi-grasp scenario and multi-object multi-grasp scenario, respectively.  The model was trained on Cornell Dataset and tested on our own multi-object dataset.  
    (b) Detection results of our system on multi-object multi-grasp scenario. The model was trained on Cornell Dataset and tested on our own multi-object dataset. Red rectangle represents the predicted grasp on each unseen object.
    (c) Experiment setting for physical grasping test. The manipulator is a 7 degree of freedom redundant robotic arm. The vision device is META-1 AR glasses with time-of-flight for RGB-D input.}
    \label{fig_case_detected}
    \label{roc_single_o_multi_g}
    \label{roc_multi_o_multi_g}
  \vspace*{-0.02in}
\end{figure*}

For realistic robotic application, a viable grasp usually depends both
on the object and its surroundings.  Given that one grasp candidate may
be impossible to achieve, there is benefit to provide a rank ordered
list of grasp candidates.  Our system provides a list of high quality
grasp candidates for a subsequent planner to select from. 
Fig. \ref{fig_example_w_label}d shows samples of the predicted grasps and
corresponding ground truths.  To evaluate the performance of the
multi-grasp detector, we employ the same scoring system as with the
single grasp, then generate the miss rate as a function of the number of false
positives per image (FPPI) by varying the detection threshold (see 
Fig. \ref{roc_single_o_multi_g}a for the single-object multi-grasp case).
The model achieves $28\%$ and $25\%$ miss rate at 1 FPPI for object-wise
split and image-wise split, respectively. 
A false positive means an incorrect grasp candidate for the object.
Thus, accepting that there may be 1 incorrect candidate grasp per image,
the system successfully detects 72\% (75\%) of possible grasps for
object-wise (image-wise) split.  The model performs slightly better in
image-wise split than object-wise split due to unseen objects in the
latter. 


%
%

\subsection{Multi-object Multi-grasp}
Here, we apply the proposed architecture to a multi-object multi-grasp
task using our Multi-Object dataset. The trained network is the same
trained network we've been reporting the results for (trained only on
the Cornell dataset with both image-split and object-split variants).
Testing involves evaluating against the multi-object dataset, and
represents a cross domain application with unseen objects. 

Fig. \ref{roc_multi_o_multi_g}a depicts the plot of miss rate versus
FPPI. At 1FPPI, the system achieves $53\%$ and $49\%$ prediction accuracy
with image-split model and object-split networks, respectively. 
Visualizations of predicted grasp candidates are depicted in Fig.
\ref{roc_single_o_multi_g}b.  The model successfully locates multiple grasp
candidates on multiple new objects in the scene with very few false
positives, and hence is practical for robotic manipulations.      


\begin {table}[t]
  \centering
  \caption {Physical Grasping Comparison \tablabel{pge} }
  \small
  \begin{tabular}{ | l | c | c | c | c |}
    \hline
    {\bf approach} & \multicolumn{2}{c|}{ \bf{Top-1}} & \multicolumn{2}{c|}{ \bf{Nearest to center}} \\ \hline
                 & \bf{detected} & \bf{physical} & \bf{detected} & \bf{physical} \\ \hline
    
    banana       & 10/10   &  7/10   & 10/10   &  8/10  \\ \hline
    glasses      &  9/10   &  8/10   &  9/10   &  9/10  \\ \hline
    ball         & 10/10   &  9/10   & 10/10   &  9/10  \\ \hline
    tape	         & 10/10   & 10/10   & 10/10   & 10/10  \\ \hline
    screwdriver  &  9/10   &  7/10   &  9/10   &  7/10  \\ \hline
    stapler      & 10/10   & 10/10   & 10/10   &  9/10  \\ \hline
    spoon        &  9/10   &  9/10   & 10/10   & 10/10  \\ \hline
    bowl         & 10/10   & 10/10   & 10/10   & 10/10  \\ \hline
    scissors     &  9/10   &  8/10   &  9/10   &  9/10  \\ \hline
    mouse        &  9/10   &  8/10   &  9/10   &  8/10  \\ \hline \hline
    average (\%) &\bf{95.0}&\bf{86.0}&\bf{96.0}&\bf{89.0}\\

    \hline
  \end{tabular}

  \vspace*{2.00ex}
  \caption {Physical Grasping Evaluation on Same Robot and Ojbects \tablabel{pge_same_robot} }
  \small
  \begin{tabular}{ | l | c | c | c |}
    \hline
    {\bf approach} &  \bf{Cornell splits} & \multicolumn{2}{c|}{ \bf{Physical grasp}} \\ \hline
                 & \bf{image / object} & \bf{detected} & \bf{physical} \\ \hline
    
    Kumra et al.	\cite{kumra2016robotic}       & 88.7   /  86.5   & 61/100   &  56/100  \\ \hline
    Guo et al. \cite{GuEtAl_ICRA2017}         & 93.8   /  89.9   & 89/100   & 81/100  \\ \hline \hline
    Ours (Top-1)                             &  96.0   /  96.1   &  95/100   &  \bf{86}/100  \\ \hline 
    Ours (center)                 &  96.0   /  96.1  &  96/100   &  \bf{89}/100  \\ 

    \hline
  \end{tabular}
  \vspace*{-1ex}
\end {table}

\subsection{Physical Grasping}
To confirm and test the grasp prediction ability in practice, a physical
grasping system is set up for experiments (see Fig. \ref{roc_multi_o_multi_g}c). 
As in \cite{watson2017real}, performance is given for both the vision
sub-system and the subsequently executed grasp movement.   The dual
scores aid in understanding sources of error for the overall experiment.
To evaluate the vision sub-system, each RGB-D input of vision sub-system
is saved to disk and annotated with the same protocol as the Cornell and
Multi-Object datasets.  The evaluation metric uses Jaccard index
$0.25$ and angle difference 30$^\circ$ thresholds. 
A set of $10$ commonly seen objects was collected from Cornell dataset for the experiment.
For each experiment, an object was randomly placed on a reachable
surface at different locations and orientations. Each object was tested
$10$ times.  The outcome of physical grasping was marked as pass or fail.

Table \tabref{pge} shows the performance of both the vision sub-system and
the physical grasping sub-system for two different policies.  For the
first (Top-1), we used the the grasp candidate with the highest
confidence score. For the second, the planner chose the grasp candidate
closest to the image-based center of the object from the top-$N$ candidates
($N=25$ in this experiment). 
In real-world physical grasping, grasp candidates close to the image-based
centroid of object should be helpful, by creating a more balanced grasp for
many objects.
The lowest performing objects
are those with a less even distribution of mass or shape (screwdriver), 
meaning that object-specific grasp priors coupled to the multi-grasp
output might improve grasping performance.  We leave this for future
work as our system does not perform object recognition. Also, the tested
objects are unseen.

For direct physical grasping comparisons, state-of-the-art 
approaches were 
implemented and applied to the same objects with the same robot. The
reference approaches selected are \cite{GuEtAl_ICRA2017} due to its
performance on the Cornell dataset, and \cite{kumra2016robotic} due to the
closely related back-bone architecture in the model design.
Table \tabref{pge_same_robot} reports both the accuracies of our 
implementations on standard Cornell dataset and physical grasping success.
The set of objects is the same as in table \tabref{pge}. 
The method in \cite{kumra2016robotic} has two parallel ResNet-50 for RGB and 
depth input, respectively.  Our implementation achieves similar results on
Cornell dataset. However, it overfits to the Cornell dataset as performance
drops substantially for real-world objects (56\% success rate).  Our
implementation of \cite{GuEtAl_ICRA2017} achieves slightly better than
reported, and reaches 81\% success rate on physical grasping. Our proposed
approach outperforms both across the board.

\begin {table}[t]
  \centering
  \caption {Physical Grasping Comparison \tablabel{pgc} }
  \small
  \begin{tabular}{ | l | c | c | c | c | c |}
    \hline
    {\bf approach} & \multicolumn{2}{c|}{ \bf{Time} (s)} & \multicolumn{2}{c|}{ \bf{Settings}}   & \bf{Success (\%)}\\ \hline
                                  & \bf{detect} & \bf{plan}  & \bf{object} & \bf{trial} & \\ \hline
    
    \cite{lenz2015deep}        &  13.50&  --    &  30   & 100   &  84 / \bf{89}* \\ \hline
    \cite{watson2017real}      &  1.80 &  --    &  10   & --    &  62     \\ \hline
    \cite{pinto2016supersizing}&  --   &  --    &  15   & 150   &  66     \\ \hline
    \cite{lu2017planning}      &  --   &2$\sim$3&  10   & --    &  84     \\ \hline
    \cite{MaEtAl_RSS[2017]}    &  0.80 &  --    &  10   & 50    &  80     \\ \hline
    \cite{MaEtAl_RSS[2017]}+refits&  2.50 &  -- &  40   & 100   &  \bf{94} \\ \hline \hline
    Ours                       &\bf{0.12}&\bf{0.10}& 10   & 100   &  \bf{89.0}\\

    \hline
  \end{tabular}
        \begin{tablenotes}
        \footnotesize
        \item[1] * Outcomes are for Baxter / PR2 robots, respectively, with the
        diffence arising from the different gripper spans.
        \end{tablenotes}
  \vspace*{-0.05in}
\end {table}

Table \tabref{pgc} further compares our experimental outcomes 
with state-of-the-art published works with physical grasping.  The testing sets for reported
experiments may include different object class/instance.  Even though
the object classes may be the same, the actual objects used could
differ.  Nevertheless, the comparison should provide some context for
grasping performance and computational costs relative to other published
approaches.
The experiments in \cite{watson2017real} had 6 objects in common with
ours.  On the common subset, \cite{watson2017real} reports a $55.0 \%$ 
success rate with a 60$s$ execution time, while ours achieves $86.7\%$
with a 15$s$ execution time (mostly a consequence of joint rate limits). 
The approach described in \cite{MaEtAl_RSS[2017]} reported $80.0 \%$
success rate on 10 household objects and $94.0 \%$ when using a cross
entropy method \cite{levine2016learning} to sample and re-fit grasp
candidates (at the cost of greater grasp detection time).   
The RL approach taking several weeks achieved $66.0 \%$ on seen and
unseen objects \cite{pinto2016supersizing}.  
Not included in the table are the reported results of
\cite{levine2016learning}, due to different experimental conditions.
They reported 90\% success on grasping objects from a bin with
replacement, and 80\% without replacement (100 trials using unseen objects).
Our approach achieves $89.0 \%$ in real-time, with subsequent planning
of the redundant manipulator taking 0.1\,secs.  Overall it exhibits
a good balance between accuracy and speed for real world object grasping
tasks.  

\subsection{Ablation Study}
This section reviews a set of experiments, summarized in Table
\tabref{abalation}, examining the contributions of the proposed
acrchitecture's components. Firstly, ResNet-50 was used to regress RGB 
input to 5D grasp configuration output (a).  This architecture can be
recognized as \cite{redmon2015real} with a deeper network and without depth
information. Then two ResNet-50 networks (b) processed RGB and depth data, 
respectively, with a small network regressing the concatenated feature for
the grasp configuration. 
This architecture matches \cite{kumra2016robotic} and boosts performance. 
However, the doubled number of parameters results in difficulties when
deploying on real-world grasping.  To keep the architecture size, one
color channel (blue) is replaced with depth information, while the performance 
is maintained (c). Next, grasp orientation is quantized and an extra branch is 
trained to classify grasping orientation of an object (d). 
The last two instances integrate grasp proposals into the ResNet-50 back-bone 
with added layers, for color (e) and RGD (f) input data. 
The multi-grasp outputs overcome averaging effects \cite{redmon2015real}
without the need to separate an image into grids. The ablation study identifies the contribution of classification, grasp proposal and the selection policy. In addition, the RGB-only version of
the proposed method is still able to achieve good performance, being
slightly worse than including depth information.

\begin {table}[t]
  \centering
  \vspace*{0.07in}
  \caption {Ablation Study \tablabel{abalation} }
  \small
  \begin{tabular}{ | l | c | c | c |}
    \hline
    \multirow{2}{*}{\centering \bf Architecture} &
    \multicolumn{2}{|c|}{\bf{Cornell Splits}} &  \bf{Number of} \\ 
    \cline{2-3}                   & \bf{image} & \bf{object} & \bf{Parameters}
    \\ \hline
    (a) RGB                       & 86.4   &  85.4     &  24559685  \\ \hline
    (b) RGB + depth               & 88.7   &  86.5     &  51738757  \\ \hline 
    (c) RGD                       & 88.1   &  86.0     &  24559685  \\ \hline 
    (d) RGD + cls*                & 89.8   &  89.3     &  24568919  \\ \hline 
    (e) RGB + cls + gp            & 94.4   &  95.5     &  28184211  \\ \hline 
    (f) RGD + cls + gp            & 96.0   &  96.1     &  28184211  \\ \hline
  \end{tabular}
    \begin{tablenotes}
        \footnotesize
        \item[1] * cls: classification; gp: grasp proposal
        \end{tablenotes}
  \vspace*{-0.05in}
\end {table}


%% file: conc.tex
We presented a novel grasping detection system to predict grasp
candidates for novel objects in RGB-D images. Compared to previous
works, our architecture is able to predict multiple candidate grasps
instead of single outcome, which shows promise to aid a subsequent grasp
planning process. 
Our regression as classification approach transforms orientation
regression to a classification task, which takes advantage of the high
classification performance of CNNs for improved grasp detection outcomes.
We evaluated our system on the Cornell grasping dataset for comparison
with state-of-the-art system using a common performance metric and
methodology to show the effectiveness of our design. We also performed
experiments on self-collected multi-object dataset for multi-object
multi-grasp scenario.   
Acceptable grasp detection rates are achieved
for the case of 1 false grasp per image.
Physical grasping experiments show a small performance loss (8.3\%)
when physically grasping the object based on correct candidate grasps found.
The outcomes might be improved by fusing the multi-grasp output with
object-specific grasp priors, which we leave to future work.
All code and data will be publicly released.